\documentclass[sn-basic,Numbered]{sn-jnl}

\usepackage{subfigure}
\usepackage{graphicx}%
\usepackage{multirow}%
\usepackage{amsmath,amssymb,amsfonts}%
\usepackage{amsthm}%
\usepackage{mathrsfs}%
\usepackage[title]{appendix}%
\usepackage{xcolor}%
\usepackage{textcomp}%
\usepackage{manyfoot}%
\usepackage{booktabs}%
\usepackage{algorithm}%
\usepackage{algpseudocode}%
\usepackage{program}%
\usepackage{listings}%
\usepackage{tabulary}%
\usepackage{array}%

\theoremstyle{thmstyleone}%
\theoremstyle{thmstyletwo}%

\theoremstyle{thmstylethree}%

\begin{document}

\title[Uncertainty measurement for complex event prediction in safety-critical systems]{Uncertainty measurement for complex event prediction in safety-critical systems}


\author*{\fnm{Maria J.} \sur{P. Peixoto}}\email{mariajoelma.pereirapeixoto@ontariotechu.net}

\author{\fnm{Akramul} \sur{Azim}}\email{akramul.azim@ontariotechu.ca}

\affil{\orgdiv{Department of Electrical, Computer and Software Engineering}, \orgname{Ontario Tech University}, \orgaddress{\city{Oshawa}, \state{Ontario}, \country{Canada}}}


\abstract{Complex events originate from other primitive events combined according to defined patterns and rules. Instead of using specialists' manual work to compose the model rules, we use machine learning (ML) to self-define these patterns and regulations based on incoming input data to produce the desired complex event. Complex events processing (CEP) uncertainty is critical for embedded and safety-critical systems. This paper exemplifies how we can measure uncertainty for the perception and prediction of events, encompassing embedded systems that can also be critical to safety. Then, we propose an approach (ML\_CP) incorporating ML and sensitivity analysis that verifies how the output varies according to each input parameter. Furthermore, our model also measures the uncertainty associated with the predicted complex event. Therefore, we use conformal prediction to build prediction intervals, as the model itself has uncertainties, and the data has noise. Also, we tested our approach with classification (binary and multi-level) and regression problems test cases. Finally, we present and discuss our results, which are very promising within our field of research and work.}



\maketitle

\section{Introduction}
An event is everything that can happen in a place at a specific time, which can be planned (interdiction schedule for road maintenance) or unplanned (a person crossing the street). For this work, we will only consider unplanned events, which can happen at any time due to the agents' actions or environmental factors, such as weather, signalling or traffic. These unexpected events can still be simple or complex. Simple or primitive events, such as vehicle speed or obstacle detection, are easily detected by sensors \cite{WANG20131808}.

Complex events combine information from different sources and can infer higher-level composite events due to patterns that arise from the analysis of simple events and even of other complex events \cite{WANG20131808}. As examples for complex events, we have congestion or accident detection and the measurement of other drivers' intentions. Besides, complex events are composed of a large amount of data from different sources that enter the event processing network to be mined. Therefore, this event processing network is responsible for house rules and patterns that satisfy complex events.

Modelling complex events processing (CEP) uncertainty is crucial as embedded and safety-critical systems widely use numerous complex events. Certainty in prediction (accuracy) is one of this system's primary requirements and concerns. This paper exemplifies how we can measure uncertainty for the perception and prediction of events, encompassing numerous embedded systems that can also be critical to safety. In addition, this work proposal can be fully applicable to several research fields, such as autonomous vehicles, health monitoring and home automation systems.

The main contributions of this paper are:
\begin{itemize}
    \item A new method capable of using machine learning to identify patterns that operate as CEP rules using a given dataset.
    \item Uncertainty recognition in primitive events using machine learning and conformal predictors to build predictions through confidence thresholds.
    \item Proposal of an approach that can be used for both classification and regression problems, generating promising results. 
\end{itemize}

In this paper, we have organized the content into different sections. First, we define the problem statement in Section \ref{sec:Problem_Statement}. Then, we describe our methodology in Section \ref{sec:Methodology}. Next, Section \ref{experiments} provides details on how we conducted tests, experiments, and evaluated the results. We also discuss related works in Section \ref{sec:RW}. Finally, we conclude our study and present future works in Section \ref{sec:Conclusion}.

\section{Problem Statement}
\label{sec:Problem_Statement}
Complex event processing systems produce events that derive from other low-level primitive events. These derived events arise from the combination of rules that fit patterns relevant to the generation of the complex event \cite{cugolaComplexEventProcessing2015}. The definition and development of these rules, which handle primitive events to produce complex events, are usually done by a specialist in the area, who defines the standards that need to be satisfied. With this, there are risks that the patterns outlined do not produce exactly the expected response or even that the rule itself contains spelling or definition errors.

In order to avoid those types of errors and speed up the process of defining rules for complex events, we propose an approach using ML. First, however, with the use of ML in the production of the rule, we need to analyze how we could explain the decisions making of our proposal and the treatment of the uncertainty associated with primitive events and propagated for complex events delivered. 

We need to measure the degree of uncertainty associated with the complex event. Thus, knowing the CEP uncertainty accurately, we may have greater control to avoid fatalities. Accordingly, our objective with this research is to use ML to compose the rules for building complex events, perform sensitivity analysis to explain the importance of inputs to the model's response and display the uncertainty linked to the complex event triggered. For the development of this paper, we will focus only on inputs from datasets.

\section{Methodology}
\label{sec:Methodology}
We define events into two main groups, primitive and derived or complex, as well as in \cite{5662592}. Primitive events are those data collected from sensors (sensors readings). Furthermore, complex events result from primitive events processing according to CEP rules (Event Patterns). 

ML models can work as complex event processors aware of the input uncertainties. In this way, the model should be responsible for elaborating the rules for triggering a complex event. With that, we would avoid human errors due to the management of the regulations. On the other hand, we will deal with a black box model in which we only know the input and output information, but we will not explicitly see the rule patterns' definitions.

Thus, we have CEP model rules $M$ that depend on time $t$ and space $s$. In addition, this model also has $n$ input parameters with uncertainties $\textbf{P} = [P_1, P_2, P_3...P_n]$ and returns an hypothesis $H$ as output, as shown below:

  \begin{equation}
  S_i(u(H)) = M(t, s, u(\textbf{P}))
  \end{equation}

Where $S_i$ is the sensitivity index generated through a quantified sensitivity analysis, and $u$ is the uncertainty quantification for each element.

Sensitivity analysis is critical because it indicates how much each uncertain parameter contributed to generating the output uncertainty. Then, we chose a variance-based sensitivity analysis to determine how much the input variation influences the output \cite{SOBOL2001271,Uncertainpy}. Thus, we use the first-order Sobol indices. These indices identify the input parameters with the most significant effect on output variability \cite{SALTELLI2002280,saltelli:2008,SALTELLI2010259}.

Sobol's first-order sensitivity index is given by: 

\begin{equation}
S_i=\frac{Var(E(H\mid\textbf{P}))}{Var(H)}
\end{equation}

Where $S_i$ is the sensitivity index, $H$ is the output resulting from the model with uncertain input parameters $\textbf{P}$. Thus, if we have a low sensitivity index, which can vary with the range [0, 1], we will have that the variance in this parameter will have little effect on the variance of the final result. Therefore, if the parameter's sensitivity index is high, any variation in this parameter will significantly affect the model's output.

The total Sobol index for parameter $P_n$ is defined according to the following equation:

  \begin{equation}
    S_{\tau i} = 1 - \frac{Var(E(H\mid P_{-n}))}{Var(H)}
  \end{equation}

We have that $Var(E(H\mid P_{-n}))$ represents the variance of the expected value of the output considering the simultaneous variation of all uncertain parameters of the set $P$ except for $P_n$. If we have $S_{\tau i}$ = 0, we say that the variability of $P_n$ has no influence on the results and can be ignored in future analyses.

Regarding uncertainty, there are two types: aleatoric and epistemic \cite{amini2020deep}. The first is related to the input data imperfections, while the last is associated with the lack of data or knowledge and errors in the prediction (model uncertainty). Aleatoric uncertainty may be more relevant when we have tasks with large volumes of data because, in this case, the epistemic uncertainty would be practically invalid due to the amount of information available. On the other hand, in smaller datasets, the quality of these data must be very high, which almost cancels out the aleatoric uncertainty. Therefore, safety-critical application data must be highly qualified for timely information processing. For this reason, we only consider epistemic uncertainty in this paper, as this impacts the analysis of the information we want.

According to the \cite{2021} and \cite{Uncertainpy} papers, there are different methods for quantifying uncertainty. Using statistical metrics \cite{2011}, the typical way is to calculate the mean or expected value and also the variance of the output. The following expression gives the mean of the output:

  \begin{equation}
  E(H) = \int_{a}^{b} hf(h)\,dh
  \end{equation}
  
Where $E(H)$ is the expected value of the output $H$, defined by the integral with range $[a,b]$,  which is the output space of $H$. Furthermore, $hf(h)\,dh$ represents the probability that $H$ is in a range width $dh$ around $h$.

The variance is defined as the expected value of the squared deviation from the mean of $H$:
    \begin{equation}\label{MSE}
        Var(H) = E((H - E(H)^2)
    \end{equation}

To generate a prediction interval of our prediction $(I_H)$, we are going to use the percentile method $(P)$ to specify the thresholds:

    \begin{equation}
    \label{conf}
        I_H = [P_{(a/2)}, P_{(1-a/2)}]
    \end{equation}
    
Therefore, we have a prediction range of 90\% of the occurrence of 90\% of all $H$ results. So 5\% of those results are below this range, and 5\% are above it. $(a/2)$ is the lower critical value and $(1-a/2)$ is the upper critical value for the standard normal distribution, and $100a\%$ is the confidence level.

Conformal predictors \cite{10.5555/1062391} is a widely used approach when we are working with uncertainties \cite{angelopoulos2021gentle, ALVARSSON202142}, being able to make predictions around confidence thresholds. Given a significance level $\alpha$, a model is capable of making predictions within a $1 - \alpha$ confidence limit. In other words, the probability of the correct label being within the predicted confidence set is almost exactly $1 - \alpha$.

\begin{figure}[htbp]
\centerline{\includegraphics[width=.6\textwidth]{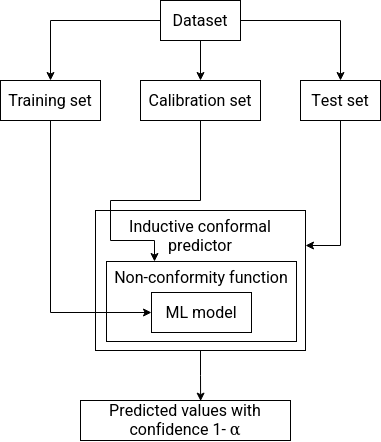}}
\caption{Conformal predictor workflow}
\label{fig:diagram3}
\end{figure}

Predictions will use ranges around the actual value in regression problems rather than predicting a single absolute value. For classification problems, the prediction will consider the possibility that each class in the data set is the proper label. Figure \ref{fig:diagram3} shows the general operation of conformal predictor work.

We use the conformal predictor shown in Figure \ref{fig:diagram3} as part of our approach proposed and explained thoroughly in Section \ref{experiments}. Therefore, according to Figure \ref{fig:diagram3}, we first divided our dataset into training, calibration and test sets. So we fit an ML model with the proper training set. Then, we define a nonconformity function that makes a sample distribution based on scores determined by the similarity of the elements in its set with those of the training set. The nonconformity function then receives the calibration set and generates the calibration scores. These scores feed the inductive conformal predictor, which is the top layer of the model. Finally, that inductive conformal predictor layer is responsible for generating predictions distributed according to the calibration scores, using as a cut-off point the significance level that varies from 0 to 1. Therefore, for example, we can say that we have an ordered set of 500 conformal scores. Then, if $\alpha = 0.1$, the cut-off point for statistical significance will be the conformal score at the 90th percentile, in this case, the 450th conformal score.

We will use in our work an inductive conformal predictor (ICP) \cite{9763489} instead of the transductive conformal predictor (TCP) \cite{pmlr-v139-cherubin21a}. We decided to focus on ICP because it is the most widely used approach to conformal predictions. Furthermore, it requires that the ML model be trained one single time for all predictions until a new amount of data is collected, and it is worth retraining to update the model \cite{10.1007/978-3-662-44722-2_28}.

\section{Experimental Design and Analysis}
\label{experiments}
For the experiments in this paper, we will use two different scenarios, a congestion identification scenario and a fire identification scenario. We will use simulated data for the first scenario and real sensor data for the second scenario.

\subsection{Scenario 1}
Congestion strongly influences mobility, accessibility and the emission of pollutants but varies widely over time and across space \cite{litman2021evaluating}. Therefore, we consider that the congestion detected event occurs when the traffic flow reaches its maximum value and the traffic density continues to increase.

\begin{figure}[htbp]
    \centerline{\includegraphics[width=.6\textwidth]{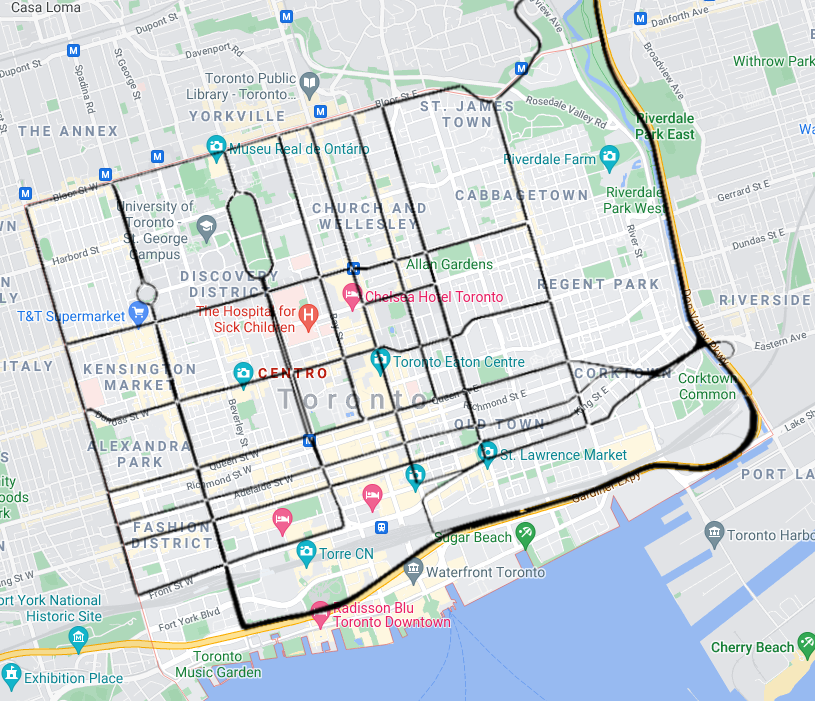}}
    \caption{Comparison of the traffic network used in the simulation with the real map of Toronto}
    \label{fig:toronto}
  \end{figure}

For this scenario, we used a traffic dataset generated through the Simulation of Urban MObility (SUMO) framework \cite{SUMO2018}. Thus, we used the net file with the Downtown Toronto abstraction made in \cite{Arasteh2021} to identify congestion points on city roads, as revealed in Figure \ref{fig:toronto}. The simulation map has 52 intersections and roads with 1 or 2 lanes. Thus, we placed SUMO lane area detectors (E2) scattered along the roads to measure the traffic generated by 1000 vehicles inserted in randomized traffic for a defined time from 0 to 100000.

We split our dataset between the training and test sets, with the test set having 25\% of the dataset's data and the training set having 75\% for experiments that did not consider uncertainty. Then we kept the same 25\% for the test dataset and split the rest evenly between the training and calibration sets in the experiments that considered uncertainty. The total number of entries in the dataset was 532,014, with the test sample size always 177,338. 

To quantify the uncertainty of our model, we use the percentile to determine what range we would like our answer to fall in. Based on equation \ref{conf}, we set this range to 97\%. For this, we set the conformal prediction significance to $\alpha = 0.03$. Thus, every time our model predicted label 0 or 1, there was a 97\% probability that this prediction was correct. 

We performed three different tests to validate our approach in scenario 1. The first experiment was a binary classification problem, the second was a multi-level classification problem, and the third was a regression problem. For the three problems, we had as input the data related to the average speed (m/s), the flow (counts), the average occupancy (\%) and the mean halting duration (s). Below we detail all the experiments.

Consider the four primitive events consumed for complex event composition:

\begin{itemize}
  \item[-] \textbf{AverageSpeed}(timestamp=1647037870, m/s=13.5)
  \item[-] \textbf{Flow} (timestamp=1647037870, counts=1.38)
  \item[-] \textbf{AverageOccupancy}(timestamp=16470, percent=3.34)
  \item[-] \textbf{HaltingDuration}(timestamp=1647037870, sec=1)
\end{itemize}

\begin{figure}[htbp]
    \centerline{\includegraphics[width=\textwidth]{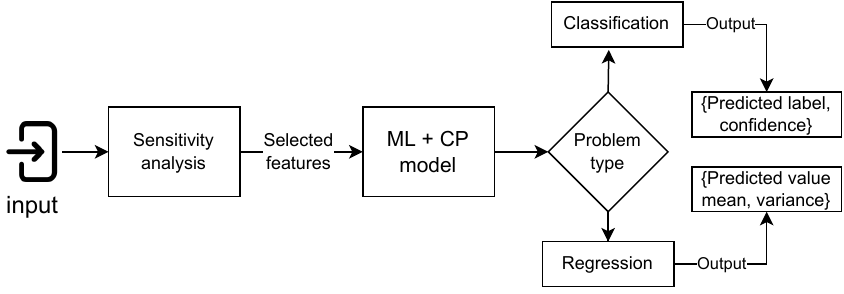}}
    \caption{Proposed approach to estimate uncertainty in events}
    \label{fig:model}
  \end{figure}

Our model presents the same structure for dealing with classification or regression problems, using a combination of Random Forest Regressor \cite{ZHANG2008618}, k-Nearest Neighbor (kNN) \cite{LIAO2002439} and conformal prediction as illustrated by Algorithm \ref{alg:model2}. First, as shown in Figure \ref{fig:model}, the model receives the inputs and verifies the essential ones through sensitivity analysis. Afterward, the selected features enter the model, composed of conformal prediction and an ML model as stated in Figure \ref{fig:diagram3}. So the prediction for the classification or regression problems can be performed. Then, our model predicts intervals from which we take the mean that indicates the prediction result.

\begin{algorithm}
\caption{Proposed ML\_CP model}\label{alg:model2}
\scriptsize
\begin{algorithmic}[1]
\Procedure{ML\_CP\_model}{$x\_train, y\_train, x\_calibrate, y\_calibrate, x\_test, problem\_type$}
    \State $underlying\_model \gets \text{RegressorAdapter}(\text{RandomForestRegressor}())$
    \State $normalizing\_model \gets \text{RegressorAdapter}(\text{KNeighborsRegressor}(n\_neighbors=5))$
    \State $normalizer \gets \text{RegressorNormalizer}(underlying\_model, normalizing\_model, \text{AbsErrorErrFunc}())$
    \State $nc \gets \text{RegressorNc}(underlying\_model, \text{AbsErrorErrFunc}(), normalizer)$
    \State $icp \gets \text{IcpRegressor}(nc)$
    \State $\text{icp.fit}(x\_train, y\_train)$
    \State $\text{icp.calibrate}(x\_calibrate, y\_calibrate)$
    \State $prediction \gets \text{icp.predict}(x\_test, significance=0.03)$
    \State $prediction\_mean \gets \text{np.mean}([prediction[:, 0], prediction[:, 1]], axis=0)$
    \State $results \gets [~]$
    \For{value in $prediction\_mean$}
        \State $frac \gets value \bmod 1$
        \State $confidence \gets 100$ if $frac < 0.1$ else $(1 - frac) * 100$ if $frac \leq 0.5$ else $frac * 100$
        \If{$problem\_type = \text{``binary classification"}$}
            \State $\text{results.append}((1$ if $value > 0.5$ else $0, confidence))$
        \ElsIf{$problem\_type = \text{``multi-classification"}$}
            \State $class\_prediction \gets \text{len}(thresholds)$
            \For{$i, threshold$ in $\text{enumerate}(thresholds)$}
                \If{$value \leq threshold$}
                    \State $class\_prediction \gets i$
                \Else
                    \State $class\_prediction \gets i + 1$
                \EndIf
            \EndFor
            \State $\text{results.append}((class\_prediction, confidence))$
        \Else
            \State $var \gets |prediction[:, 1] - prediction[:, 0]|$
            \State $\text{results.append}((value, var))$
        \EndIf
    \EndFor
    \State \Return $results$
\EndProcedure
\end{algorithmic}
\end{algorithm}

For comparison, we implemented two other replicable approaches per the paper's information and definitions \cite{s21051863}. Thus, Algorithms \ref{alg:alg1} and \ref{alg:alg2} represent the Normal Probabilistic Model (NPM) and Improved Probabilistic Model (IPM) approaches for classification (Algorithm \ref{alg:alg1}) and regression (Algorithm \ref{alg:alg2}) problems. The difference between NPM and IPM declared by the \cite{s21051863} authors is the probability distribution function of the error, which in NPM is N(0, 1) and in IPM is the normal distribution of the standard deviation representing the inaccuracy of sensors specified by the manufacturer.

In scenario 1, we are working with simulated data, and therefore, we need information regarding the factory inaccuracy of the sensors. Considering this, we assume that sensor uncertainty is proportional to the standard deviation of the simulated data ($stds * 0.1$). In this way, we have the probability distribution function of the error equal to $[\mathcal{N}(0, 1), \mathcal{N}(0, 1), \mathcal{N}(0, 1), \mathcal{N}(0, 1) )]$ for NPM and $[\mathcal{N}(0, 0.7), \mathcal{N}(0, 0.4), \mathcal{N}(0, 1.0), \mathcal{N}(0, 3.7)]$ for IPM relative to the four input primitive events.

\begin{algorithm}
\caption{Probabilistic Model Classification}\label{alg:alg1}
\scriptsize
\textbf{Class} ProbabilisticModelClassification
\begin{algorithmic}[1]
\Function{\_\_init\_\_}{$self, detector\_stds$}
    \State $self.means \gets None$
    \State $self.stds \gets detector\_stds$
\EndFunction

\Function{Fit}{$self, X, y$}
    \State $n\_features \gets length(X[1])$
    \State $n\_classes \gets length(unique(y))$
    \State $self.means \gets zero\_matrix(n\_classes, n\_features)$

    \For{$label$ in $range(n\_classes)$}
        \State $X\_label \gets X[y == label]$
        \State $self.means[label] \gets mean(X\_label)$
    \EndFor
\EndFunction

\Function{Predict}{$self, X$}
    \State $n\_classes \gets length(self.means)$
    \State $log\_likelihoods \gets zero\_matrix(length(X), n\_classes)$

    \For{$label$ in $range(n\_classes)$}
        \State $log\_likelihoods[:, label] \gets$
        \Statex $sum(norm\_logpdf(X, self.means[label], self.stds))$
    \EndFor

    \State \Return $argmax(log\_likelihoods)$
\EndFunction
\end{algorithmic}
\end{algorithm}

\begin{algorithm}
\caption{Probabilistic Model Regression}\label{alg:alg2}
\scriptsize
\textbf{Class} ProbabilisticModelRegression
\begin{algorithmic}[1]
\Function{\_\_init\_\_}{$self, feature\_stds$}
    \State $self.means \gets None$
    \State $self.feature\_stds \gets feature\_stds$
\EndFunction

\Function{Fit}{$self, X, y$}
    \State $n\_features \gets length(X[1])$
    \State $self.means \gets zero\_array(n\_features)$

    \For{$i$ in $range(n\_features)$}
        \State $self.means[i] \gets mean(y * X[:, i])$
    \EndFor
\EndFunction

\Function{Predict}{$self, X$}
    \State $y\_pred \gets zero\_array(length(X))$

    \For{$i$ in $range(length(X))$}
        \State $weights \gets 1 / (self.feature\_stds ^ 2)$
        \State $y\_pred[i] \gets sum(X[i, :] * self.means * weights) / sum(X[i, :] * weights)$
    \EndFor

    \State \Return $y\_pred$
\EndFunction
\end{algorithmic}
\end{algorithm}

\textbf{Analysis of the proposed method for a binary classification problem:}

The first experiment is a simple binary classification problem. The objective is to determine whether there is congestion (no-congestion=0, congestion=1).

\begin{figure}[htbp]
    \centerline{\includegraphics[width=\textwidth]{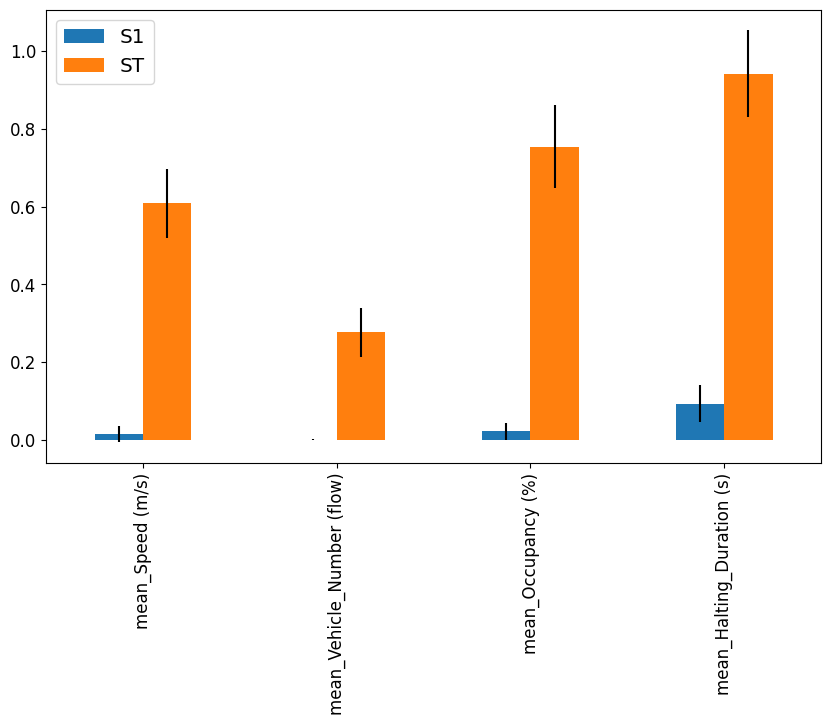}}
    \caption{Sobol analysis to indicate priority input for binary classification problem}
    \label{fig:sens_bin_class}
  \end{figure}

With our proposed approach, the sensitivity analysis indicates ``speed", ``occupancy" and ``mean\_Halting\_Duration (s)" as the essential inputs (Figure \ref{fig:sens_bin_class}). S1 means first-order Sobol indices, and ST indicates total Sobol indices. First-order Sobol indices consider each parameter's direct contribution, excluding interaction terms, to a specific output. The total Sobol indices provide a view of all interactions of a given input, not providing details of which parameter this input interacts with nor in which order.

Following the Algorithm \ref{alg:model2}, our approach initially predicts a range between 0 and 1, for which we calculate a mean. Then, the estimated mean value is used as a confidence index to indicate the choice of a label. For example, if we have a mean between 0 and 0.50, the output label will be 0, and the confidence index will range from 50\% to 100\%, depending on the mean. On the other hand, if the mean is from 0.51 to 1, the chosen label will be 1, with a confidence index ranging from 51\% to 100\%. Consequently, if we have $mean=0.30,~then~label=0~and~confidence=70\%$.

We calculated the sensitivity and specificity of each model as defined by Equations \ref{sensitivity} and \ref{specificity}, respectively.

\begin{equation}\label{sensitivity} 
    Sensitivity = \frac{TP}{TP + FN}
\end{equation}

\begin{equation}\label{specificity} 
    Specificity = \frac{TN}{TN + FP}
\end{equation}

\begin{enumerate}
  \item[TP] - the number of true positives.
  \item[FN] - the number of false negatives.
  \item[TN] - the number of true negatives.
  \item[FP] - the number of false positives.
\end{enumerate}

After the previous clarifications, we present the obtained results in Table \ref{table:binary_results}.

\begin{table}[!ht]
\caption{Models performance for the binary classification problem}
\label{table:binary_results}
\begin{tabular}{|l|l|l|l|}
\hline
& \textbf{ML\_CP} & NPM & IPM \\ \hline
Accuracy & \textbf{0.98} & 0.64 & 0.63\\ \hline
Sensitivity & \textbf{0.97} & 0.31 & 0.29 \\ \hline
Specificity & 0.98 & \textbf{0.99} & \textbf{0.99} \\ \hline
\end{tabular}
\end{table}

We analyzed the accuracy, sensitivity and specificity of each model. The sensitivity indicates the proportion of label 1 classes that were actually classified as label 1 by the model (true positive). On the other hand, specificity indicates the proportion of label 0 that was classified as label 0 (true negative). Thus, considering sensitivity and specificity, in addition to accuracy, the ML\_CP model presents the best results. Although NPM and IPM have a marginally higher specificity rate than ML\_CP, it does not necessarily mean that the NPM and IPM models were effective. The NPM and IPM models tended to classify most of their predictions as 0, resulting in high specificity but low sensitivity and accuracy.

\begin{figure}[htbp]
    \centerline{\includegraphics[width=\textwidth]{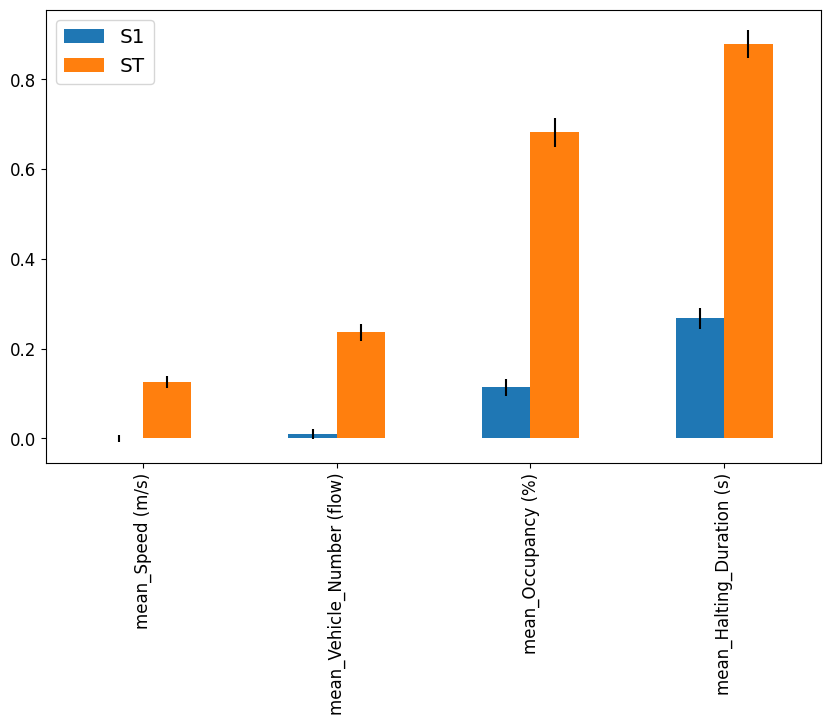}}
    \caption{Sobol analysis to indicate priority input for multi-level classification problem}
    \label{fig:sens_mult_class}
  \end{figure}

\textbf{Analysis of the proposed method for a multi-level classification problem:}
For the experiment involving the multi-classification problem, we replaced the last column of our dataset, which initially only indicated whether there was (label 1) or not (label 0) congestion. In place of this last column, we now have three different situations in case there is congestion: light congestion (label 1), moderate congestion (label 2) and severe congestion (label 3), in addition to label 0 in case there is no congestion. Also, we maintain the same four analysis features (speed, flow, occupancy and halting\_duration). For the ML\_CP model, we only use the features selected by sensitivity analysis presented in Figure \ref{fig:sens_mult_class}, which means ``mean\_Vehicle\_Number (flow)", ``mean\_Occupancy (\%)" and ``mean\_Halting\_Duration (s)".

The traffic jam level was defined based on the average jam length in meters. Thus, when the average congestion length is less than or equal to 82.5 meters, we have congestion level 1. When the congestion length is more significant than 82.5 meters and less than or equal to 165 meters, we define congestion as level 2. Finally, if the congestion length exceeds 165 meters, we have congestion level 3.

Table \ref{table:Multi-level} presents the performance of the models for this problem, quantifying the global model sensitivity and specificity, as defined in Equations \ref{sensitivity} and \ref{specificity}, respectively. In addition, the table also discriminates the accuracy of each of the models.

\begin{table}[!ht]
\caption{Models performance for the multi-level classification problem}
\label{table:Multi-level}
\begin{tabular}{|l|l|l|l|}
\hline
& \textbf{ML\_CP} & NPM & IPM \\ \hline
Accuracy & \textbf{0.98} & 0.72 & 0.74\\ \hline
Sensitivity & \textbf{0.97} & 0.71 & 0.74 \\ \hline
Specificity & \textbf{0.97} & 0.75 & 0.76 \\ \hline
\end{tabular}
\end{table}

 The models can classify their results with labels 0, 1, 2 or 3. As shown in Table \ref{table:Multi-level}, the results for the NPM and IPM models are pretty similar. The ML\_CP model still has more promising results.

\textbf{Analysis of the proposed method for a regression problem}:

We use the same dataset used in the other two experiments for this regression problem, changing only the target column. Thus, we have speed, flow, occupancy and halting duration as default data features. In addition, we defined a mean\_max\_jam\_length\_in\_meters column as the target, ranging from zero to 250 meters. For this problem, the model would have to predict the max value of the jam length considering the provided features.

\begin{figure}[htbp]
    \centerline{\includegraphics[width=\textwidth]{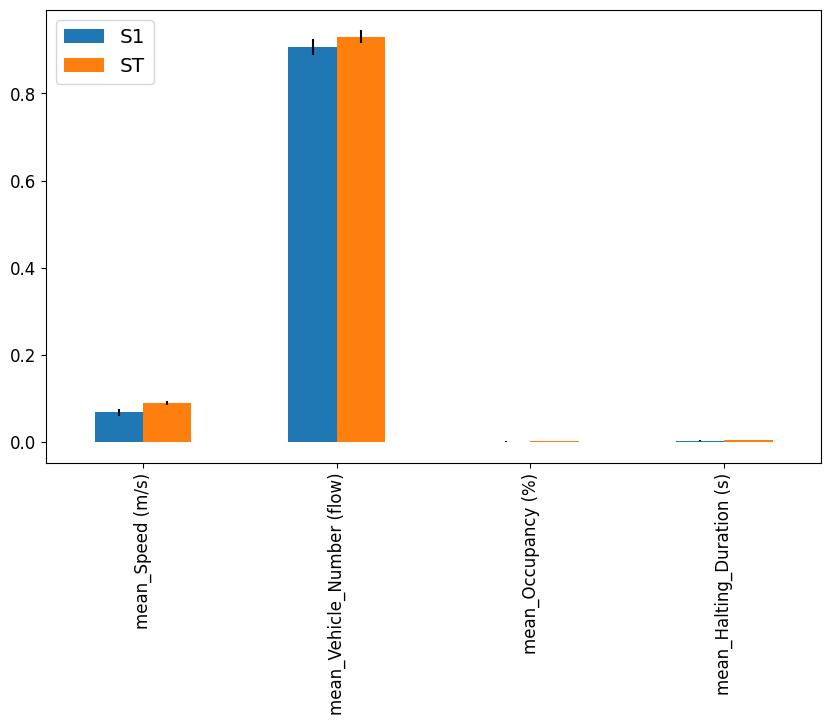}}
    \caption{Sobol analysis to indicate priority input for regression problem}
    \label{fig:s_regression}
  \end{figure}

For the ML\_CP model, we used our proposal discussed above (Algorithm \ref{alg:model2}). The inputs selected through sensitivity analysis for this problem are ``mean\_Speed (m/s)" and ``mean\_Vehicle\_Number (flow)", displayed in Figure \ref{fig:s_regression}. Thus, we calculated an interval with a 97\% chance of containing the value in meters of the congestion size, which can vary from 0 to 250 meters. Then, we calculate the mean and the variance based on the interval values found, which are our model outputs for the specified regression problem.

The performance of regression models was evaluated, and the results are presented in Table \ref{table:regression}. The metrics used for evaluation were $R^2$, mean absolute error, mean squared error and median absolute error. The results of Table \ref{table:regression} indicate that our ML\_CP proposal outperformed other models in this regression problem, as well as in the classification problems previously analyzed.

\begin{table}[!ht]
\caption{Models performance for the regression problem}
\label{table:regression}
\begin{tabular}{|l|l|l|l|}
\hline
& \textbf{ML\_CP} & NPM & IPM \\ \hline
$R^2$ & \textbf{0.99} & -132.0 & -8.11\\ \hline
Mean absolute error & \textbf{0.29} & 234.39 & 59.4 \\ \hline
Mean squared error & \textbf{2.76} & 124276.16 & 8512.38 \\ \hline
Median absolute error & \textbf{0.003} & 118.93 & 36.07 \\ \hline
\end{tabular}
\end{table}

\subsection{Scenario 2}
This scenario involves detecting fires in various environments by analyzing temperature, smoke, and flame characteristics. The dataset used in this study was collected and supplied by the authors of \cite{Umoh}. To build this dataset, the researchers used three sensors: a DHT11 for measuring temperature, an MQ-2 smoke sensor, and an LM393 flame sensor. This dataset was also used to assess the effectiveness of the proposed DST-CEP \cite{s21051863}. Thus, we reproduce the tests to evaluate and compare our approach (ML\_CP) with the DST-CEP, Normal Probabilistic Model (NPM) and Improved Probabilistic Model (IPM).

\begin{figure}[htbp]
    \centerline{\includegraphics[width=\textwidth]{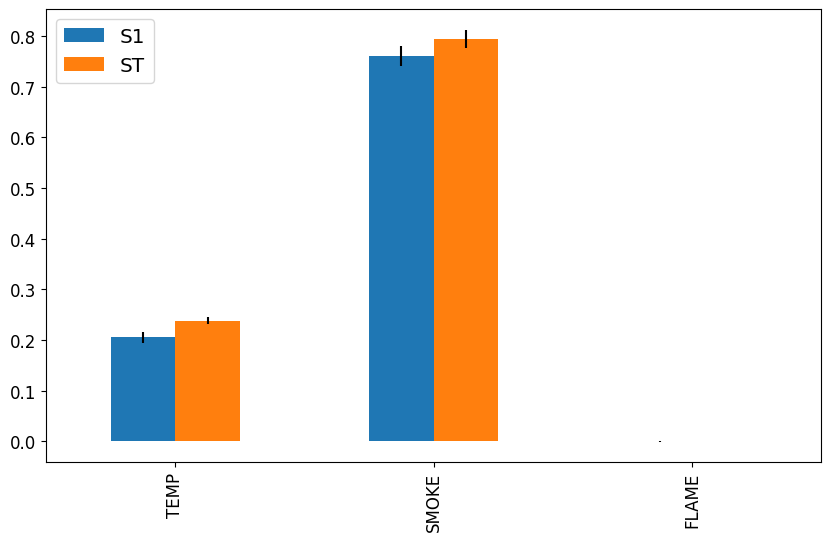}}
    \caption{Sobol analysis to indicate priority input for fire detection problem}
    \label{fig:si_fire_outbreak}
  \end{figure}

In the sensitivity analysis of our model, shown in Figure \ref{fig:si_fire_outbreak}, only the temperature and smoke features were considered to be of high relevance for the detection (label 1) or non-detection (label 0) of fire. Consequently, as input, we only use the temperature and smoke values.

In Table \ref{table:comp_RW2}, we have presented the results of our performance analysis compared to three other approaches. The table clearly shows that our solution surpasses all others in terms of Accuracy, Precision, Recall, and F-Measure metrics.

\begin{table}[!ht]
\caption{Models performance for fire detection scenario}
\label{table:comp_RW2}
\footnotesize
\begin{tabular}{|l|l|l|l|l|}
\hline
 Approach & Accuracy & Precision & Recall & F-Measure \\ \hline
 NPM & 81.14\% & 67.32\% & 55.68\% & 60.95\% \\ \hline
 IPM & 81.43\% & 68.21\% & 55.68\% & 61.31\% \\ \hline
 DST-CEP & 95.00\% & 85.71\% & 97.30\% & 91.14\% \\ \hline
 \textbf{ML\_CP} & \textbf{99.00\%} & \textbf{98.00\%} & \textbf{99.00\%} & \textbf{99.00\%} \\ \hline
\end{tabular}
\end{table}

\section{Related Work}
\label{sec:RW}
The number of works considering the uncertainty in producing their results is increasing. The paper "Deep Evidential Regression" \cite{amini2020deep} synthesizes two main types of uncertainties, aleatoric and epistemic. Aleatoric uncertainty is called perturbation (data uncertainty), and epistemic uncertainty is called imprecision (model uncertainty). These two types of uncertainties are independent and can also be combined.

The work \cite{s21051863} investigates the identification and treatment of uncertainty in CEP-based IoT applications. For this, the authors use Dempster–Shafer Theory to propose their approach - DST-CEP. According to this paper, the commonly observed CEP uncertainty lies in primitive events (i.e., sensor readings) and rules that compose complex events (i.e., high-level situations). Thus, the study proposes to combine data from unreliable sensors in conflicting situations and detect correct results from the uncertainty measurement.

The authors of ``Complex Event Processing Over Uncertain Data" \cite{10.1145/1385989.1386022} present a method that considers the production of new events in the face of uncertainty. They have extended the complex event processing literature to manage data with uncertainty. However, to our understanding, the authors of the work \cite{10.1145/1385989.1386022} do not present the probability (which would indicate the degree of uncertainty as defined in \cite{troya_uncertainty_2021}) of the new event produced occurring in the final result.

In the paper ``Complex Event Processing over Uncertain Data Streams" \cite{5662592}, the authors work with one processing engine extension to support the complex events processing in uncertain data. The research also defined how the input and output data should be. In addition, the researchers calculated and presented the confidence probability in the complex event's output. In our opinion, the definition of what the input data should look like limits the data that can be used as input to produce complex events. Another difference in our approach to that work is that is no analysis in \cite{5662592} of how much input data contribute to the output uncertainty.

Uncertainty measurement is also necessary for event representation languages. In \cite{2011}, the authors declare a probabilistic logic programming language (ProbLog) to represent uncertainty using probabilistic estimates as independent random variables. Similar to our proposal, the authors of the research \cite{2011} also use upper and lower bounds to calculate the success probability. A confidence interval is essential for estimating uncertainty in events, with wide intervals indicating high uncertainty, or short intervals, indicating low uncertainty.

Images and videos are typical for composing a complex event and are generally from different sources. Therefore, those data sources are usually loaded with uncertainties, especially when collected in general conditions, such as low lighting, background noise and occlusion. Thus, works like \cite{6890893} deal with the uncertainty arising from those data for the composition of complex events using human activities recognition, validated by matching what was produced with actual observations. Furthermore, the authors' approach \cite{6890893} takes into account the uncertainty generated by the input and how reliable the output can be, with the classification of something they call a belief degree.  

Uncertainty is present in the most varied solutions for the Internet of Things. It means that the sensors can be noisy, the data transmission can be affected by the available network or even the rules of pattern combinations to trigger a complex event can contain faults. The paper \cite{9509992} states precisely this need to face the challenges for producing self-adaptive approaches to the context~\cite{peixoto2023improving}. Due to the high demand for smart objects, the measurement and presentation of uncertainty can guarantee the quality of services.

Comparing our proposal with the related works (see Table \ref{table:comp_RW}), we consider the following evaluation points:

\begin{enumerate}
\item[1] - Complex event analysis
\item[2] - Uncertainty measurement
\item[3] - ML for complex events rules establishment
\item[4] - Use of Sobol indices to explain outputs based on inputs
\item[5] - The proposal cover classification and regression problems
\item[6] - Uncertainty measurement using conformal prediction
\item[7] - Use of common metrics to measure uncertainty in other areas, such as in clinical tests
\end{enumerate}

\begin{table}[!ht]
\caption{Comparison between related works and our approach}
\label{table:comp_RW}
\centering
\begin{tabular}{cccccccc}
\toprule
Papers\textbackslash{}Points & 1 & 2 & 3 & 4 & 5 & 6 & 7 \\
\midrule
\cite{amini2020deep} & \checkmark & \checkmark & X & X & X & X & X \\
\cite{10.1145/1385989.1386022} & \checkmark & \checkmark & X & X & X & X & Just sensitivity \\
\cite{5662592} & \checkmark & \checkmark & X & X & X & X & Just sensitivity \\
\cite{2011} & \checkmark & \checkmark & X & X & \checkmark & X & X \\
\cite{6890893} & \checkmark & \checkmark & \checkmark & X & X & X & X \\
\cite{9509992} & \checkmark & \checkmark & \checkmark & X & \checkmark & X & X \\
ML\_CP & \checkmark & \checkmark & \checkmark & \checkmark & \checkmark & \checkmark & \checkmark (Sensitivity and specificity) \\
\bottomrule
\end{tabular}
\end{table}

Table \ref{table:comp_RW} shows a comparison of our approach with the related works mentioned above. According to Table \ref{table:comp_RW}, our research is the only one that covers all the contribution points of this study, at least until the moment of this writing.

\section{Conclusion}
\label{sec:Conclusion}
This work proposes a model that considers ML, sensitivity analysis and uncertainty measurements for complex events composition for classification and regression problems. Supporting a forecast with a stated confidence level is crucial, as we constantly deal with insufficient or inaccurate data and the model's uncertainty. Thus, conformal prediction is a lean approach for developing prediction intervals.

We have found that using conformal prediction with machine learning models and sensitivity analysis allows for more accurate predictions of potential failures. Our ML\_CP model performed the best in all of our experiments. However, we must test this approach in different situations to fully understand its advantages and disadvantages for predictive modelling. In the future, we plan to explore how we can incorporate other types of data, such as camera footage, social media, and sensor data, into our proposal. We also intend to use the complex events and their associated uncertainty as inputs for reinforcement learning agents, such as autonomous vehicles~\cite{peixoto2021using, peixoto2023improving}.

\section*{Declarations}
\subsection*{Ethical Approval}
Not applicable.

\subsection*{Competing interests}
The authors have no financial or proprietary interests in any material discussed in this article.

\subsection*{Authors' contributions}
Both authors (Maria J. P. Peixoto and Akramul Azim) contributed and revised the manuscript equally.

\subsection*{Funding}
No funding was received for conducting this study.

\subsection*{Availability of data and materials}
Any data from this paper is available from the corresponding author upon request.

\bibliography{sn-article}

\end{document}